\documentclass[letterpaper, 10 pt, journal, twoside]{ieeetran} 

\IEEEoverridecommandlockouts                              

\usepackage{booktabs}

\usepackage[noadjust]{cite}
\nocite{*}  

\makeatletter
\let\NAT@parse\undefined
\makeatother

\usepackage{hyperref}
\usepackage{subcaption}
\usepackage{pdfpages}
\usepackage{balance}
\usepackage{tabularx}
\usepackage{adjustbox}
\usepackage{comment}
\hypersetup{
    colorlinks=true,
    citecolor=[RGB]{0,0,0},
    linkcolor=[RGB]{0,0,0},
    filecolor=[RGB]{0,0,0},      
    urlcolor=[RGB]{0,0,0},
    pdfpagemode=FullScreen,
}
\usepackage{graphicx}
\usepackage{hyperref}
\usepackage{keyval}
\usepackage{algorithm}
\usepackage{algpseudocode}
\usepackage{amsmath, amsfonts}
\usepackage{amssymb}
\usepackage{cleveref}
\usepackage{fancyhdr} 
\usepackage{color}
\usepackage{transparent}
\usepackage{booktabs}
\usepackage{tabularx}

\usepackage{caption}
\usepackage{subcaption}
\usepackage{textcomp}
\usepackage{stfloats}
\usepackage{url}
\usepackage{verbatim}
\usepackage{graphicx}
\usepackage{multirow}
\usepackage{mathtools}
\let\labelindent\relax
\usepackage{enumitem} 
\usepackage{booktabs}

\usepackage{caption}
\usepackage{tabularx}
\usepackage{threeparttable}
\usepackage{makecell}

\newcommand{\idest}{i.e., }
\newcommand{\exempli}{e.g., }

\newif\ifshowstatshorizontal
\showstatshorizontaltrue

\usepackage{xargs} 
\usepackage[colorinlistoftodos,prependcaption,textsize=tiny]{todonotes}
\newcommandx{\unsure}[2][1=]{\todo[linecolor=red,backgroundcolor=red!25,bordercolor=red,#1]{#2}}
\newcommandx{\change}[2][1=]{\todo[linecolor=blue,backgroundcolor=blue!25,bordercolor=blue,#1]{#2}}
\newcommandx{\info}[2][1=]{\todo[linecolor=OliveGreen,backgroundcolor=OliveGreen!25,bordercolor=OliveGreen,#1]{#2}}
\newcommandx{\improvement}[2][1=]{\todo[linecolor=Plum,backgroundcolor=Plum!25,bordercolor=Plum,#1]{#2}}
\newcommandx{\thiswillnotshow}[2][1=]{\todo[disable,#1]{#2}}
 
\def\HiLiYellow{\leavevmode\rlap{\hbox to \hsize{\color{yellow!20}\leaders\hrule height .8\baselineskip depth .4ex\hfill}}}
  
\fancypagestyle{firststyle}
{
  \fancyhf{}

}

\fancypagestyle{secondstyle}
{
  \fancyhf{}
  \fancyhead[R]{\footnotesize \thepage}

}

\usepackage{cleveref}

\graphicspath{{figures/}}

\newcommand{\includegraphicsdpi}[3]{
    \pdfimageresolution=#1  
    \includegraphics[#2]{#3}
    \pdfimageresolution=20  
}

\title{
SEAL: Towards Safe Autonomous Driving via Skill-Enabled \\ Adversary Learning for Closed-Loop Scenario Generation
}

\author{Benjamin Stoler$^{1}$ \and Ingrid Navarro$^{1}$ \and Jonathan Francis$^{1,2}$ \and Jean Oh$^{1}$
\thanks{Manuscript received: February, 17, 2025; Revised May, 30, 2025; Accepted June, 28, 2025.}
\thanks{This paper was recommended for publication by Editor Ashis Banerjee upon evaluation of the Associate Editor and Reviewers' comments.
This work was supported by the Korean Ministry of
Trade, Industry, and Energy (MOTIE; grant \#P0026022),
and by the Korea Institute of Advancement of Technology
(KIAT), through the International Cooperative R\&D program
(\#P0019782): Embedded AI Based fully autonomous driving
software and Maas technology development.} 
\thanks{$^{1}$Benjamin Stoler, Ingrid Navarro, and Jean Oh are with the School of Computer Science, Carnegie Mellon University
        {\tt\footnotesize \{bstoler, ingridn, jeanoh\}@cs.cmu.edu}}%
\thanks{$^{2} $Jonathan Francis is with the Bosch Center for Artificial Intelligence and also with the School of Computer Science, Carnegie Mellon University {\tt\footnotesize jon.francis@us.bosch.com}}%

\thanks{Digital Object Identifier (DOI): see top of this page.}
}

\usepackage{xstring}
\usepackage{xparse}
\usepackage{etoolbox}

\newcommand{\rebuttalClean}{}

\makeatletter
\ifdefined\rebuttalClean
  \let\origtextcolor\textcolor

  \def\textcolor#1#2{%
    \def\colname{#1}%
    \def\red   {red}%
    \def\purple{purple}%
    \def\blue  {blue}%
    \ifx\colname\red
    \else
      \ifx\colname\purple
        \origtextcolor{black}{#2}%
      \else
        \ifx\colname\blue
          \origtextcolor{black}{#2}%
        \else
          \origtextcolor{#1}{#2}%
        \fi
      \fi
    \fi
  }%
\fi
\makeatother

\begin{document}

\definecolor{egoblue}{HTML}{00B0F0}
\definecolor{advred}{HTML}{CD0000}

\maketitle


\renewcommand{\thefootnote}{\fnsymbol{footnote}}

\begin{abstract}

Verification and validation of autonomous driving (AD) systems and components is of increasing importance, as such technology increases in real-world prevalence. Safety-critical scenario generation is a key approach to robustify AD policies through closed-loop training. However, existing approaches for scenario generation rely on simplistic objectives, resulting in overly-aggressive or non-reactive adversarial behaviors. To generate diverse adversarial yet realistic scenarios, we propose SEAL, a scenario perturbation approach that leverages learned objective functions and adversarial, human-like skills. SEAL-perturbed scenarios are more realistic than SOTA baselines, leading to improved ego task success across real-world, in-distribution, and out-of-distribution scenarios, of more than 20\%. To facilitate future research, we release our code and tools: \url{https://navars.xyz/seal/}

\end{abstract}


\markboth{IEEE Robotics and Automation Letters. Preprint Version. Accepted June, 2025}
{Stoler \MakeLowercase{\textit{et al.}}: SEAL: Towards Safe Autonomous Driving via Skill-Enabled Adversary Learning}

\section{Introduction} \label{sec:introduction}


\begin{figure*}[t!]
\begin{center}
   \includegraphicsdpi{200}{width=0.90\textwidth}{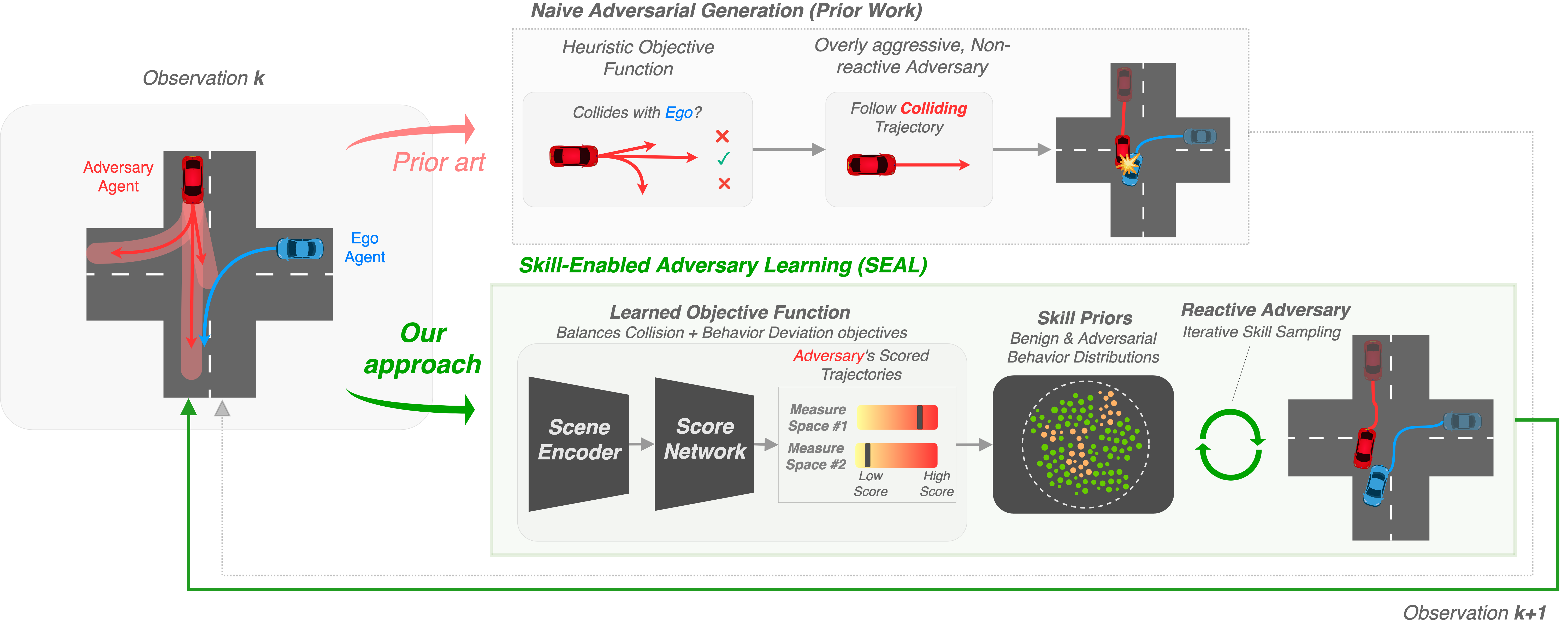}
   \vspace{-0.2cm}
   \captionof{figure}{An overview of SEAL. Our scenario generation approach leverages a learned objective function and an adversarial skill-based, reactive policy, for improved adversary realism and more effective closed-loop training, leading to safer autonomous driving agents, compared to previous approaches such as CAT~\cite{zhang2023cat} and GOOSE~\cite{ransiek2024goose}.} %
   \label{fig:overview}
   \vspace{-0.3cm}
\end{center}
\end{figure*}


\IEEEPARstart{W}{ith} the growing deployment of autonomous driving (AD) technologies in real-world settings, ensuring the safety of such systems has only increased in importance and public concern~\cite{cummings2024assessing, xing2022influences}. As AD verification and validation approaches continue to evolve, scenario-based testing via datasets and simulation has emerged as a core methodology, where alternatives such as on-road testing via a sufficiently large number of miles driven can be prohibitively expensive, risky, and infeasible~\cite{kalra2016driving, lou2022testing}. While validation of system behavior under normal operating circumstances is valuable, testing AD behavior under \textit{safety-critical} and other corner-case circumstances is vital for Safety of the Intended Functionality (SOTIF) standards~\cite{ISO21448, song2024industry, zhang2022finding}.

Scenarios are often curated in the form of large datasets of real-world recorded driving traces, providing a basis for assessing human behaviors and for training machine learning models~\cite{ettinger2021large, wilson2023argoverse, caesar2021nuplan}. AD subsystems are then asked to perform tasks such as forecasting the future motion of various road users or controlling the behavior of certain vehicles in a simulated reconstruction~\cite{gu2021densetnt, suo2023mixsim, montali2024waymo, park2020diverse}. However, the presence of critical scenarios in collected datasets is exceedingly low, a problem identified as the ``curse-of-rarity'' in autonomous driving~\cite{feng2023dense, stoler2024safeshift, liu2024curse}. Thus, programmatically \textit{generating} safety-critical scenarios is necessary. To ensure that generated scenarios retain realistic properties, it is appealing to perturb the behavior of one or more agents in a principled way, rather than using first principles to painstakingly assemble a scenario from scratch~\cite{ding2023survey, huang2024cadre, cao2022advdo, zhang2023cat}.  
In this setting, one agent is referred to as the \textit{ego} agent, while the modified background traffic participants are \textit{adversary} agent(s), who attempt to attack the ego in some way.

State-of-the-art (SOTA) approaches in perturbation-based scenario generation have coupled a dynamic scenario generation framework with an ego control policy being trained with closed-loop objectives~\cite{zhang2023cat, yang2024drivearena, tian2024enhancing}, in contrast with previous less-efficient staged approaches~\cite{rempe2022generating, xu2023diffscene}. These approaches can still be sub-optimal, however, in that they can struggle to provide \textit{useful} training stimuli to a closed-loop agent. In particular, we identify three key issues in recent SOTAs: 1) they have a limited view of safety-criticality, \exempli focusing only on inducing collisions or near-misses; 2) they lack reactivity to an ego agent's behavior diversity; and 3) their optimization objectives tend to maximize ``unrealistic'' and overly-aggressive adversarial behavior, limiting their usefulness for balanced model training. 

Therefore, in this paper, we propose and evaluate a method for \textbf{S}kill-\textbf{E}nabled \textbf{A}dversary \textbf{L}earning (SEAL), which yields significantly improved downstream ego behavior, in closed-loop training with safety-critical scenario generation. 
Our method addresses the identified limitations in prior art by introducing two novel components, as shown in \Cref{fig:overview}. %
First, we introduce a learned objective function to \textit{anticipate} how a reactive ego agent will respond to a candidate adversarial agent behavior. We quantify both collision closeness and induced ego behavior deviation, thus providing a broadened understanding of safety criticality.
Second, we develop a skill-enabled, reactive adversary policy; in particular, inspired by human cognition, we leverage a hierarchical framework that is akin to how humans operate vehicles~\cite{medeiros2014hierarchical} and we create an adversarial prior that selects human-like \textit{skill primitives} to increase criticality while maintaining realism.

Furthermore, we argue that safety-critical scenario generation should be evaluated based on behavior realism and usefulness for ego policy improvement, not just induced criticality. Prior work often assesses ego policies on generated scenarios where safety-critical behavior remains effectively \textit{in-distribution} with respect to training data and heuristic perturbations~\cite{zhang2023cat, hanselmann2022king}.
To address this, we build on recent scenario characterization work, SafeShift~\cite{stoler2024safeshift}, to identify real (non-generated) but safety-relevant scenarios, enabling a more realistic, out-of-distribution evaluation. While in-distribution performance is informative, real-world performance on challenging \textcolor{red}{\textbf{[Reviewer 2, Item 1]: }}\textcolor{purple}{scenarios} is ultimately most important.

In summary, our paper comprises three main contributions:
\begin{enumerate}
    \item We propose two novel techniques for safety-critical perturbation: (i) a learned objective function to select candidate trajectories; and (ii) an adversarial skill-based, reactive policy for more realism in adversary behavior.
    \item We design an improved evaluation setting for closed-loop training, utilizing real-world safety-relevant scenarios in contrast to just in-distribution generated \textcolor{purple}{scenarios}. 
    \item We provide results on several key experiments, showing an increase of more than 20\% in ego task success rate over SOTA baselines, across scenarios generated closed-loop by our proposed framework, across scenarios generated closed-loop by previous SOTA baseline frameworks, \textit{and} across real-world safety-relevant scenarios. 
\end{enumerate}

\section{Related Work} \label{sec:related_works}

\subsection{Scenario Generation in Autonomous Driving}

Approaches for generating scenarios that reproduce the distribution of \textit{normal} driving behavior have been extensively explored. Some methods ensure the diversity of generated traffic behavior~\cite{xu2023bits, suo2021trafficsim}, while others aim for controllability through rule-based or language-driven specifications~\cite{lu2024scenecontrol, zhong2023guided, zhong2023language}. However, due to the rarity of safety-critical events in recorded data~\cite{feng2023dense, stoler2024safeshift, liu2024curse}, other approaches have focused on directly generating corner-case scenarios by injecting adversarial behaviors. Earlier works in safety-critical scenario generation relied on gradient-based optimization approaches with access to vehicle dynamics \cite{hanselmann2022king, rempe2022generating, wang2021advsim}, a limitation in model-free settings. Other methods, such as diffusion-based approaches~\cite{xu2023diffscene, chang2024safesim}, are compute-intensive and impractical to be used in a closed-loop manner. Efficient methods like CAT~\cite{zhang2023cat} and GOOSE~\cite{ransiek2024goose}, which leverage trajectory prediction priors and reinforcement learning (RL) respectively, prioritize simple collision objectives and are non-reactive to the ego agent.
\textcolor{red}{\textbf{[Reviewer 4, Item 4]:}} \textcolor{purple}{Similarly,~\cite{zhang2024learning} employs reactive adversaries but focuses only on collisions for criticality and defines realism via proximity to ground-truth trajectories, making it sensitive to distribution shifts.}
\textcolor{blue}{In contrast, our approach efficiently generates reactive, nuanced adversarial behavior across multiple axes of criticality, providing a stronger closed-loop training signal.}

\subsection{Robust Training and Evaluation in Autonomous Driving}

Several techniques for robustifying AD policies against safety-critical and out-of-distribution scenarios have been explored. Formal methods, such as Hamilton-Jacobi (HJ) reachability, have been utilized in various driving tasks, but struggle with dimension scaling~\cite{chen2021safe, qin2022quantifying}. Similarly, domain randomization has been used as a form of data augmentation (\exempli randomizing vehicle control parameters~\cite{voogd2023reinforcement} or scenario initial states~\cite{huang2024safetydr}) but requires excessive sampling to cover a sufficient domain size. Thus, adaptive stress testing~\cite{feng2023dense, koren2018adaptive} and adversarial training have been increasingly used, either as a fine-tuning scheme~\cite{xu2023diffscene, rempe2022generating} or in a fully closed-loop training pipeline~\cite{zhang2023cat, wang2021advsim}, providing continuous feedback to an ego agent. However, these approaches still tend to optimize for naive collision objectives alone.

Evaluation of robust training and scenario generation approaches is crucial. Many works evaluate generated scenarios against fixed rule-based or replay ego planners alone~\cite{chang2024safesim, ransiek2024goose, suo2023mixsim, rempe2022generating, ding2020learning}, offering limited insights into the efficacy of adversarial agents against more sophisticated ego agents. Additionally, adversarially-trained ego policies are often tested on scenarios perturbed by the same adversarial method used in training~\cite{zhang2023cat, hanselmann2022king, xu2023diffscene, wang2021advsim}, leading to in-distribution evaluations. Conversely, we focus on out-of-distribution evaluation of well-trained, reactive ego policies, in both adversarial scenarios perturbed by \textit{other} SOTA approaches, as well as real safety-relevant scenarios.

Out-of-distribution evaluation has been well-explored in AD trajectory prediction~\cite{stoler2024safeshift, ye2023improving, itkina2023interpretable, park2020diverse}, but these approaches often aim to characterize an entire \textcolor{purple}{scenario} without focusing on a single ego driver or identifying a specific adversary. In AD control tasks, some prior work has explored out-of-distribution settings, such as CARNOVEL~\cite{filos2020can, francis2022distribution}, which tests unseen scenario types like roundabouts. Additionally, Lu et al.~\cite{lu2023imitation} evaluate across real-world scenarios of various difficulty levels, but do not hold out the hardest \textcolor{purple}{scenarios} during training. Our approach thus addresses this gap by offering a more comprehensive and rigorous evaluation, across a wide set of adversarial and real-world scenarios.

\section{Preliminaries} \label{sec:preliminaries}

In this section, we define relevant notation and task definitions used in the rest of this paper. Let $(x,\ y)^{(t)}$ represent the location of an agent (\idest vehicle, pedestrian, or cyclist) in the ground plane at some given time $t$. We then define an agent's trajectory as the ordered set $X = ((x\ ,y)^{(t)}\ |\ t \in \{1,2,...,T\})$ over $T$ timesteps at some fixed time delta. 

\noindent\textbf{Base Scenario:} We define a base scenario, $\mathbf{S}$, as the tuple $(\mathbf{X},\ \mathbf{M},\ \texttt{ego},\ \texttt{adv})$, with $\mathbf{X} = \{X_{i}\ |\ i \in \{1,2,...,N\}\}$ consisting of the set of all agent trajectories observed, where $X_{i}$ denotes the trajectory of an agent with the ID of $i$, and $N$ is the total number of agents. All relevant map and scenario meta information (such as lane connectivity, traffic light locations, etc.) is given as $\mathbf{M}$. Finally, \texttt{ego} and \texttt{adv} refer respectively to the agent IDs of the ego vehicle (to be controlled in simulation) and the adversarial vehicle (to be perturbed to induce criticality).

\noindent\textbf{Scenario Perturbation Task:} For this task, $K$ re-simulations of a base scenario $\mathbf{S}$ are performed as episodes, where agents start from the same state as the base scenario and follow a behavior prescribed by some policy (\idest a reactive policy or predefined trajectory), which may be different than their original trajectory. Let $\widetilde{\mathbf{X}}^{(k)}$ represent the observed trajectories in the $k$-th re-simulation of $\mathbf{S}$. The perturbation-based safety critical scenario generation task is thus assigning behaviors to roll-out for all non-ego agents, conditioned on the base scenario $\mathbf{S}$ and $K$ previous episodes, $\{\widetilde{\mathbf{X}}^{(k)}\ |\ k \in \{1,2,...,K\}\}$, such that the resulting $\widetilde{\mathbf{X}}^{(K + 1)}$ satisfies some specified desired properties of criticality.
Importantly, we treat the ego agent's behavior as a \textit{black box}: while we are able to observe previous behavior as $\widetilde{\mathbf{X}}^K_{\text{ego}}$, we have no access to the model or any privileged information on \texttt{ego}'s decision-making process. \textcolor{red}{\textbf{[Reviewer 1, Item 4]: }}\textcolor{purple}{In practice, during training, we maintain a queue of the most recent $K$ perturbation roll-outs for each base scenario; during evaluation, we instead run $K$ sequential perturbation–simulation steps and use the final roll-out as the adversarial scenario.}

\section{Approach: Skill-Enabled Adversary Learning for Scenario Generation} \label{sec:approach}

\textcolor{blue}{To increase scenario criticality while preserving realism, we propose the \textbf{S}kill-\textbf{E}nabled \textbf{A}dversary \textbf{L}earning (SEAL) approach for perturbation-based scenario generation.}
\textcolor{red}{\textbf{[Reviewer 1, Item 6]: }}\textcolor{purple}{Similar to CAT~\cite{zhang2023cat}, SEAL employs a probabilistic trajectory predictor $\pi_{\text{gen}}$ to sample candidate adversary futures conditioned on $\mathbf{S}$. However, directly executing these samples has three main limitations: 1) it measures criticality only via collisions, ignoring, \exempli forced ego hard braking or swerving; 2) it prevents reactivity to ego decisions; and 3) it often generates non-human-like behavior, driving straight at the ego with no avoidance. SEAL addresses these with a learned objective for flexible trajectory selection and an adversarial skill policy for more human-like and reactive behavior.}

\begin{figure}[t]
    \centering
    \includegraphics[width=0.8\linewidth]{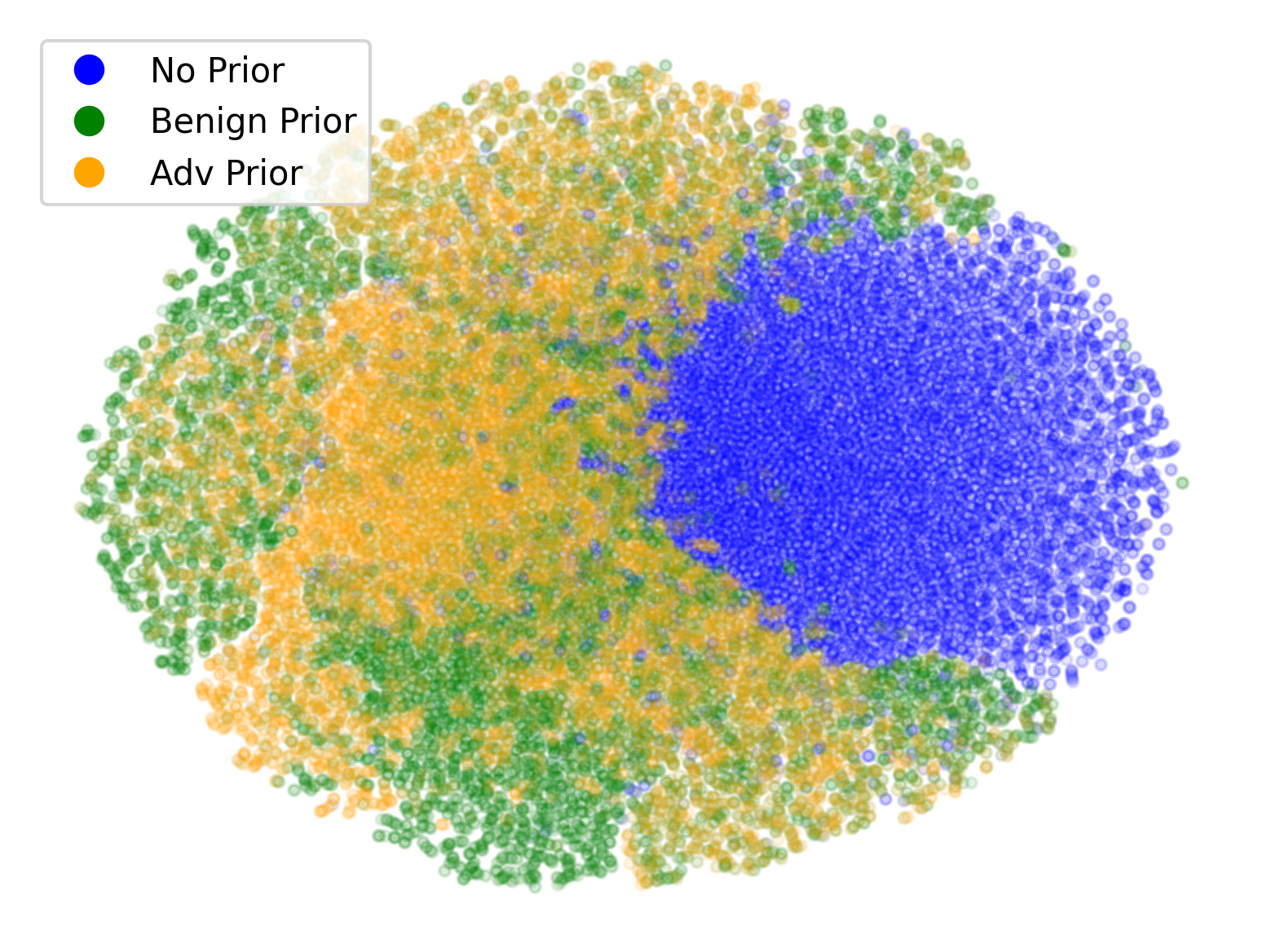}
    \vspace{-0.4cm}
    \caption{Skill space visualized with t-SNE~\cite{van2008visualizing}. Benign and adversarial priors map to several regions representing useful, human-like skills, with meaningful separation and overlap.}
    \label{fig:skills}
    \vspace{-0.3cm}
\end{figure}

\subsection{Learned Objective Function} \label{ssec:learned_obj}

Many previous works rely on heuristic approaches to select the \textit{best} trajectory from a candidate set to be assigned to the behavior of the adversary agent, $\widetilde{X}^{(K+1)}_{\text{adv}}$. For instance, CAT~\cite{zhang2023cat} compares bounding box overlaps across the previous $K$ episodes in all candidate routes, selecting the one which collides with the most previous ego roll-outs at the earliest time step or is closest to a collision, otherwise. We instead aim to select among candidate trajectories in a more flexible way that captures both closeness to collision as well as likelihood of anticipated ego behavior deviation (\exempli causing the ego to swerve or execute a hard-brake maneuver). 

We frame the problem as a supervised regression task. First, we build a dataset of simulated outcomes, where we roll out and observe all trajectory pairs of ego and adversarial agents, ($\widetilde{X}^{(K+1)}_{\text{ego}}$, $\widetilde{X}^{(K+1)}_{\text{adv}}$). To keep ego behavior as a black-box in downstream closed-loop training, we have the ego follow a reactive heuristic policy during this stage. 
We then obtain ground truth values from the collected demonstrations, using the following scoring functions, 
similar to measure functions used in prior work~\cite{stoler2024safeshift, huang2024cadre}: %
\begin{equation}\label{eq:col}
    f_{\text{coll}} = 
        \text{exp} \left(-\frac{1}{b}\min\limits_{t} \left|\left| \widetilde{X}^{(k),t}_\text{ego} - \widetilde{X}^{(k),t}_\text{adv} \right|\right|_2 \right)
\end{equation}
\begin{equation}\label{eq:diff}
    f_{\text{diff}} = 1 - \text{exp} \left(-\frac{1}{b}\sum_{t} \left|\left| \widetilde{X}^{(k - 1),t}_\text{ego} - \widetilde{X}^{(k),t}_\text{ego} \right|\right|_2\right),  
\end{equation}
\noindent where $b \in \mathbb{R}$ is a hyperparameter controlling sensitivity to distance values. Both \Cref{eq:col} and \Cref{eq:diff} map to $[0, 1]$, where $1$ indicates maximal criticality and $0$ indicates minimal. \Cref{eq:col} captures collision closeness between the ego and adversary over a given roll-out, while \Cref{eq:diff} captures ego behavior difference between two episodes. However, instead of only assessing past episodes, we propose to \textit{predict} these measures for a roll-out yet to happen by training a neural network, $\pi_{\text{score}}$ (detailed in \Cref{ssec:approach_impl}). This $\pi_{\text{score}}$ network aims to predict $f_{\text{coll}}$ and $f_{\text{diff}}$ conditioned on a previous $\widetilde{X}^{(k)}_{\text{ego}}$ and the proposed $\widetilde{X}^{(K+1)}_{\text{adv}}$. The final score for ranking candidate trajectories is the sum of the predicted $f_\text{coll}$ and $f_\text{diff}$ values from $\pi_\text{score}$, averaged over the $K$ previous ego roll-outs.
\textcolor{red}{\textbf{[Reviewer 1, Item 2]: }}\textcolor{purple}{By using $\pi_\text{score}$ in place of additional heuristic simulation, we enable scoring to be conditioned on actual ego policy roll-outs and substantially reduce runtime overhead.}

\subsection{Adversarial Skill Learning}\label{ssec:adv_skill}

We design a reactive policy $\pi_{\text{adv}}$, to guide the adversary's behavior, unlike recent works \cite{zhang2023cat, ransiek2024goose}, where the selected adversary follows a predefined trajectory. This adversarial policy observes and acts in a closed-loop simulator alongside the ego policy. In this context, skill-based hierarchical policies are appealing approaches as they capture maneuvers at a higher abstraction, compared to the low-level actions of a simulator, corresponding more closely to how humans operate vehicles~\cite{medeiros2014hierarchical}. 

We build upon prior work, ReSkill~\cite{rana2023residual}, which utilizes expert demonstrations to extract paired observation and action sequences as state-conditioned ``skills'' which are then embedded using a Variational AutoEncoder (VAE). Additionally, a state-conditioned prior network is trained to map from a state to a useful location in the VAE's latent space to be decoded into a reconstructed skill for the agent to follow. 

In our work, we separate the demonstrated skills into adversarial (\idest those ending in a collision or near-miss) and benign skills (\idest those avoiding a collision while staying on road).
\textcolor{red}{\textbf{[Reviewer 4, Item 1]:}} \textcolor{purple}{We use a sliding-window partitioning scheme that excludes segments starting within twice the skill horizon before an out-of-road event, and labels as adversarial those within the same window before a collision. This reflects the intuition that not only the final skill but also preceding behavior contributes to unsafe outcomes.}
We then train two prior networks in parallel with a shared-skill VAE: benign skills flow through a ``benign'' prior while adversarial skills flow through an analogous ``adversarial'' prior.
\textcolor{red}{\textbf{[Reviewer 4, Item 2]:}} \textcolor{purple}{In this way, the adversarial agent policy, $\pi_{\text{adv}}$, leverages the adversarial prior to select skills likely to lead to safety-critical outcomes. Furthermore, because each prior is implemented as a real-valued non-volume preserving transformation trained on observed data (as in ReSkill), sampled noise vectors bijectively correspond to plausible, in-distribution behaviors.}
\Cref{fig:skills} visualizes the learned skill spaces over uniformly sampled states; regions of overlap correspond to skills which may be useful to both an adversarial and benign agent (\exempli lane-keeping, smooth kinematics, etc.) while distinct regions correspond to skills only useful for that particular agent (\exempli for an adversary: cutting-off another vehicle, hard-braking in a dangerous way, etc.). 

To integrate this skill module with the trajectory generation and ranking discussed in \Cref{ssec:learned_obj}, we first select the highest ranking candidate trajectory, $\widetilde{X}^{(K+1)}_{\text{adv}}$. We derive goals and subgoals from this selected trajectory to provide to $\pi_{\text{adv}}$ as navigation information. Skills are then executed in a hierarchical manner as in \cite{rana2023residual}: at the start of the episode or when a skill has completed, a new skill is selected based on the current observation and adversarial prior. The agent then decodes that skill, in a closed-loop manner, into raw actions. To further increase safety-criticality, the adversary initially exactly follows $\widetilde{X}^{(K+1)}_{\text{adv}}$ before switching to this adversarial skill policy at a fixed offset before the anticipated point of maximal collision risk.

\subsection{SEAL Implementation Details}\label{ssec:approach_impl}

For training and validating both the learned objective function and skill spaces, we leverage the well-established Waymo Open Motion Dataset (WOMD)~\cite{ettinger2021large} dataset, as well as a subset of \textcolor{purple}{scenarios} therein labeled by Waymo as containing interacting agents. A further subset of 500 of these \textcolor{purple}{scenarios} has been used by prior work, and we henceforth refer to this set as \texttt{WOMD-Normal}~\cite{zhang2023cat, huang2024versatile}. We split these \textcolor{purple}{scenarios} into 400 training and 100 evaluation \textcolor{blue}{examples}.

For $\pi_{\text{gen}}$, we utilize a pre-trained DenseTNT~\cite{gu2021densetnt} trajectory prediction model, as used by CAT. \textcolor{red}{\textbf{[Reviewer 1, Item 1]: }}\textcolor{purple}{$\pi_{\text{gen}}$ takes as input the first one second of $\textbf{X}$, as well as the static meta information $\textbf{M}$, and produces 32 candidate eight-second future adversary paths.} We use the MetaDrive simulator~\cite{li2022metadrive} and its included IDM policy~\cite{treiber2000congested} as the heuristic reactive agent to collect imperfect demonstration data, described and utilized in both \Cref{ssec:learned_obj} and \Cref{ssec:adv_skill}. For data augmentation, \textbf{all} agents in the \textcolor{purple}{scenario} follow the IDM policy and produce useful demonstrations, rather than collecting examples from solely the ego. \textcolor{red}{\textbf{[Reviewer 1, Item 3]: }}\textcolor{purple}{We extract subgoals from each trajectory using MetaDrive’s default waypoint logic, placing checkpoints every 8 meters as navigation input for $\pi_\text{adv}$.} 

We implement $\pi_{\text{score}}$ as a VectorNet-style polyline encoder~\cite{gao2020vectornet}, followed by a multilayer perceptron decoder to the predicted values of $f_{\text{coll}}$ and $f_{\text{diff}}$. We use an MSE loss objective on the sum of the two values, ensuring equal weight to both predicted measures. For $\pi_{\text{adv}}$, we leverage the skill embedding framework from \cite{rana2023residual}, with identical architectures and loss functions across our two parallel prior networks. We empirically set the hyperparameter $b$ in \Cref{eq:col} and \Cref{eq:diff} to 8, use a skill time horizon of 10 steps, and fix $K$ to 5 (consistent with CAT).

\section{Experimental Setup} \label{sec:experimental_setup}

\begin{table*}[hbtp]
\vspace{0.5em}
\centering
\caption{Ego performance on adversarially-perturbed (a, b, c) and unmodified, real-world (d, e) scenarios. \texttt{WOMD-Normal} are WOMD~\cite{ettinger2021large} \textcolor{purple}{scenarios} with basic interactive agents labeled by Waymo; \texttt{WOMD-SafeShift-Hard} refers to SafeShift-mined~\cite{stoler2024safeshift} real \textcolor{purple}{scenarios} in WOMD. Adversarially-perturbed \textcolor{purple}{scenarios} use \texttt{WOMD-Normal} as base scenarios, in both training and evaluation settings. \textit{Higher} success rates and lower crash and out of road rates are better. 
\textcolor{red}{\textbf{[Reviewer 2, Item 2]: }}\textcolor{purple}{Ego realism scores are shown in (f), averaged over settings (a--e) using Wasserstein distance (WD); lower is better.}
}
\vspace{-0.3cm}
\label{tab:results_main}
\begin{tabular}{ccc}
    \begin{subtable}{0.31\textwidth}
        \centering
        \caption{WOMD-Normal, GOOSE-Gen~\cite{ransiek2024goose}}
        \begin{adjustbox}{max width=\linewidth}
        \begin{tabular}{p{1.7cm}p{1.3cm}p{1.3cm}p{1.5cm}}
        \toprule
        \textbf{Training} & \textbf{Success} & \textbf{Crash} & \textbf{Out of Road} \\
        \midrule
        \textit{None} (Replay) & 0.59 (0.00) & 0.41 (0.00) & 0.00 (0.00) \\
        \cline{1-4}
        No Adv & 0.41 (0.06) & 0.37 (0.02) & 0.23 (0.04) \\
        GOOSE & 0.37 (0.07) & 0.35 (0.09) & 0.30 (0.17) \\
        CAT & 0.35 (0.03) & 0.27 (0.02) & 0.39 (0.06) \\
        \textbf{SEAL} & \textbf{0.44} (0.04) & 0.27 (0.00) & 0.27 (0.00) \\
        \bottomrule
        \end{tabular}
        \end{adjustbox}
    \end{subtable}
    &
    \begin{subtable}{0.31\textwidth}
        \centering
        \caption{WOMD-Normal, CAT-Gen~\cite{zhang2023cat}}
        \begin{adjustbox}{max width=\linewidth}
        \begin{tabular}{p{1.7cm}p{1.3cm}p{1.3cm}p{1.5cm}}
        \toprule
        \textbf{Training} & \textbf{Success} & \textbf{Crash} & \textbf{Out of Road} \\
        \midrule
        \textit{None} (Replay) & 0.18 (0.00) & 0.82 (0.00) & 0.00 (0.00) \\
        \cline{1-4}
        No Adv & 0.32 (0.01) & 0.46 (0.02) & 0.22 (0.01) \\
        GOOSE & 0.25 (0.10) & 0.47 (0.02) & 0.31 (0.04) \\
        CAT & 0.32 (0.03) & 0.32 (0.03) & 0.40 (0.00) \\
        \textbf{SEAL} & \textbf{0.42} (0.02) & 0.32 (0.04) & 0.24 (0.02) \\
        \bottomrule
        \end{tabular}
        \end{adjustbox}
    \end{subtable}
    &
    \begin{subtable}{0.31\textwidth}
        \centering
        \caption{WOMD-Normal, \textbf{SEAL}-Gen}
        \begin{adjustbox}{max width=\linewidth}
        \begin{tabular}{p{1.7cm}p{1.3cm}p{1.3cm}p{1.5cm}}
        \toprule
        \textbf{Training} & \textbf{Success} & \textbf{Crash} & \textbf{Out of Road} \\
        \midrule
        \textit{None} (Replay) & 0.32 (0.00) & 0.68 (0.00) & 0.00 (0.00) \\
        \cline{1-4}
        No Adv & 0.33 (0.03) & 0.50 (0.05) & 0.21 (0.04) \\
        GOOSE & 0.26 (0.08) & 0.46 (0.00) & 0.27 (0.06) \\
        CAT & 0.31 (0.00) & 0.34 (0.04) & 0.36 (0.02) \\
        \textbf{SEAL} & \textbf{0.38} (0.04) & 0.36 (0.01) & 0.25 (0.06) \\
        \bottomrule
        \end{tabular}
        \end{adjustbox}
    \end{subtable}
\end{tabular}
\begin{tabular}{ccc}
   \vspace{0.8em} 
    \begin{subtable}{0.31\textwidth}
        \centering
        \caption{WOMD-Normal, Real \textcolor{purple}{Scenarios}}
        \begin{adjustbox}{max width=\linewidth}
        \begin{tabular}{p{1.7cm}p{1.3cm}p{1.3cm}p{1.5cm}}
        \toprule
        \textbf{Training} & \textbf{Success} & \textbf{Crash} & \textbf{Out of Road} \\
        \midrule
        \textit{None} (Replay) & 1.00 (0.00) & 0.00 (0.00) & 0.00 (0.00) \\
        \cline{1-4}
        No Adv & 0.48 (0.02) & 0.21 (0.01) & 0.28 (0.04) \\
        GOOSE & 0.44 (0.13) & 0.23 (0.03) & 0.34 (0.10) \\
        CAT & 0.50 (0.02) & 0.15 (0.06) & 0.36 (0.10) \\
        \textbf{SEAL} & \textbf{0.59} (0.01) & 0.15 (0.00) & 0.27 (0.01) \\
        \bottomrule
        \end{tabular}
        \end{adjustbox}
    \end{subtable}
    &
    \begin{subtable}{0.31\textwidth}
        \centering
        \caption{WOMD-SafeShift-Hard, Real \textcolor{purple}{Scenarios}}
        \begin{adjustbox}{max width=\linewidth}
        \begin{tabular}{p{1.7cm}p{1.3cm}p{1.3cm}p{1.5cm}}
        \toprule
        \textbf{Training} & \textbf{Success} & \textbf{Crash} & \textbf{Out of Road} \\
        \midrule
        \textit{None} (Replay) & 0.97 (0.00) & 0.01 (0.00) & 0.02 (0.00) \\
        \cline{1-4}
        No Adv & 0.28 (0.05) & 0.38 (0.05) & 0.33 (0.02) \\
        GOOSE & 0.19 (0.04) & 0.42 (0.06) & 0.36 (0.04) \\
        CAT & 0.24 (0.00) & 0.38 (0.03) & 0.37 (0.05) \\
        \textbf{SEAL} & \textbf{0.38} (0.02) & 0.29 (0.02) & 0.33 (0.04) \\
        \bottomrule
        \end{tabular}
        \end{adjustbox}
    \end{subtable}
    &
    \begin{subtable}{0.32\textwidth}
        \centering
        \caption{\textcolor{red}{\textbf{[Reviewer 2, Item 2]:}}\textcolor{purple}{Aggregate Realism}}
        \begin{adjustbox}{max width=\linewidth}
        \begin{tabular}{p{1.7cm}p{1.3cm}p{1.3cm}p{1.7cm}}
        \toprule
        \textcolor{purple}{\textbf{Training}} & \textcolor{purple}{\textbf{Yaw WD}} & \textcolor{purple}{\textbf{Acc WD}} & \textcolor{purple}{\textbf{Road WD}} \\
        \midrule
        \textcolor{purple}{\textit{None} (Replay)} & \textcolor{purple}{0.014} & \textcolor{purple}{0.269} & \textcolor{purple}{0.004} \\
        \cline{1-4}
        \textcolor{purple}{No Adv} & \textcolor{purple}{0.147} & \textcolor{purple}{\textbf{3.041}} & \textcolor{purple}{\textbf{0.252}} \\
        \textcolor{purple}{GOOSE} & \textcolor{purple}{0.152} & \textcolor{purple}{3.052} & \textcolor{purple}{0.312} \\
        \textcolor{purple}{CAT} & \textcolor{purple}{0.154} & \textcolor{purple}{3.050} & \textcolor{purple}{0.374} \\
        \textcolor{purple}{\textbf{SEAL}} & \textcolor{purple}{\textbf{0.146}} & \textcolor{purple}{3.074} & \textcolor{purple}{0.270} \\
        \bottomrule
        \end{tabular}
        \end{adjustbox}
    \end{subtable}
\end{tabular}
\vspace{-0.5cm}
\end{table*}

\begin{table*}[hbtp]
\vspace{0.9em}
\centering
\caption{Scenario generation quality; results are averaged over all tested ego
models. WD measures are Wasserstein distances over adversary behavior; a lower
value indicates greater realism. \textcolor{red}{\textbf{[Reviewer 4, Item 5]: }}\textcolor{purple}{Lower collision velocities (m/s) and head-on rates are better.} A \textit{lower} ego \texttt{Success} is better, as this table assesses safety-critical effectiveness.}
\label{tab:gen_quality}
\resizebox{0.99\textwidth}{!}{\begin{tabular}{lllllllll}
\toprule
Eval Scenario Type & Ego Success ($\downarrow$) & Realism WD ($\downarrow$) & Yaw WD ($\downarrow$) & Acc WD ($\downarrow$) & Road WD ($\downarrow$) & \textcolor{purple}{Coll. Vel. ($\downarrow$)} & \textcolor{purple}{Head-On ($\downarrow$)} & \textcolor{purple}{Head-On (Severe) ($\downarrow$)} \\
\midrule
WOMD-Normal,
Real \textcolor{purple}{Scenarios} & 60.0\% & \textcolor{blue}{0.056} & \textcolor{blue}{0.120} & \textcolor{blue}{0.020} & \textcolor{blue}{0.027} & \textcolor{purple}{2.849} & \textcolor{purple}{06.7\%} & \textcolor{purple}{04.2\%} \\
WOMD-SafeShift-Hard,
Real \textcolor{purple}{Scenarios} & 41.3\% & \textcolor{blue}{0.069} & \textcolor{blue}{0.116} & \textcolor{blue}{0.044} & \textcolor{blue}{0.043} & \textcolor{purple}{2.206} & \textcolor{purple}{00.0\%} & \textcolor{purple}{00.0\%} \\
\midrule
WOMD-Normal,
GOOSE-Gen & 43.0\% & \textcolor{blue}{0.401} & \textcolor{blue}{0.124} & \textcolor{blue}{0.601} & \textcolor{blue}{0.482} & \textcolor{purple}{4.744} & \textcolor{purple}{13.2\%} & \textcolor{purple}{11.7\%} \\
WOMD-Normal,
CAT-Gen & \textbf{29.6\%} & \textcolor{blue}{0.167} & \textcolor{blue}{0.123} & \textcolor{blue}{0.305} & \textcolor{blue}{0.074} & \textcolor{purple}{4.136} & \textcolor{purple}{\textbf{07.1\%}} & \textcolor{purple}{05.9\%} \\
WOMD-Normal,
\textbf{SEAL}-Gen & 31.9\% & \textcolor{blue}{\textbf{0.108}} & \textcolor{blue}{\textbf{0.121}} & \textcolor{blue}{\textbf{0.157}} & \textcolor{blue}{\textbf{0.049}} & \textcolor{purple}{\textbf{2.950}} & \textcolor{purple}{09.2\%} & \textcolor{purple}{\textbf{03.6\%}} \\
\bottomrule
\end{tabular}
}
\vspace{-0.5cm}
\end{table*}

We leverage SEAL to generate \textcolor{purple}{scenarios} for two primary purposes: providing data augmentation during closed-loop training of reinforcement learning (RL) agent policies, and providing a means of evaluating such agents' capabilities.

\subsection{Policy Training} \label{ssec:policies}

For closed-loop training of an ego agent policy, we leverage the \texttt{WOMD-Normal} set
along with the MetaDrive simulator~\cite{li2022metadrive}, described in \Cref{ssec:approach_impl}. Then, we follow the curriculum training approach proposed by CAT~\cite{zhang2023cat}, where a random base scenario $\mathbf{S}$ from the train split is selected and has a random chance of being perturbed; this perturbation chance increases throughout the training process. Agents observe the environment via simulated LiDAR returns and navigation information based on their original destination in $\mathbf{X}$. Agents act on the environment with normalized steering and acceleration forces as $\mathbf{a}$; the ego and adversarial agents follow either a policy or a predefined trajectory, while all other agents follow their original trajectory in $\mathbf{X}$. 

We utilize ReSkill~\cite{rana2023residual} as our underlying RL algorithm, a recent SOTA approach in hierarchical RL.
We use our skill space built in \Cref{ssec:adv_skill}, utilizing the benign prior rather than the adversarial one. \textcolor{blue}{The low-level action learned by the ReSkill agent is a remediating $\Delta \mathbf{a}$ adjustment to the action decoded based on the current skill and state pair, $\mathbf{a'}$, while the high-level action selects the noise vector to be passed to the prior.} Thus, the action sent to the environment is $\mathbf{a} = \mathbf{a'} + \Delta \mathbf{a}$. Actions are performed at a 10Hz rate,
and all agents are trained for one million timesteps in total, empirically sufficient for consistent policy convergence.

\subsection{Evaluation Settings} \label{ssec:datasets}

Many previous works evaluate agent performance, in-distribution, on a held-out subset of their \textit{own} generated \textcolor{purple}{scenarios} \cite{zhang2023cat, hanselmann2022king, xu2023diffscene, wang2021advsim}. For additional comprehensiveness, we propose to utilize a recent scenario characterization approach, SafeShift~\cite{stoler2024safeshift}, for identifying real-world safety-relevant base scenarios, denoted as \texttt{WOMD-SafeShift-Hard}. We start by identifying scenarios containing interacting agents labeled by Waymo. We then apply SafeShift's hierarchical scoring to these agents and select scenarios where the \textit{interacting} agents have trajectory scores in the top 20th percentile across WOMD, randomly sampling 100 \textcolor{purple}{scenarios} therein. The ego and adversary agents are assigned to the interacting agents with the higher and lower trajectory score, respectively.

We baseline SEAL against two recent SOTA safety critical scenario generation approaches, that can be utilized in a closed-loop manner: CAT~\cite{zhang2023cat} and GOOSE~\cite{ransiek2024goose}. CAT heuristically chooses a trajectory from $\pi_{\text{gen}}$ to apply to the adversarial agent; we use the same $\pi_{\text{gen}}$ function for both CAT and SEAL, for fairness.
GOOSE learns to iteratively modify control points of a NURBS~\cite{ma1998nurbs} curve fit to the original adversary's trajectory, observing the outcome of each roll-out.
We train GOOSE against the MetaDrive IDM agent using the \texttt{WOMD-Normal} training set and GOOSE's ``deceleration'' task goal---induce a collision while maintaining kinematic feasibility. For consistency, we limit the number of GOOSE policy steps (\idest observed roll-outs) to $K=5$. 

\subsection{Metrics} \label{ssec:metrics}

Within MetaDrive, episodes are terminated when the ego agent either arrives safely at its goal (\texttt{Success}), collides with another agent (\texttt{Crash}), or violates an off-road constraint (\idest crosses a road edge or yellow median; \texttt{Out of Road}). As such, we report these corresponding rates as the key metrics for ego performance, following prior work~\cite{zhang2023cat}.

For evaluating generated scenario quality, we examine the induced ego \texttt{Success} rate, across all tested ego methods. We derive a realism metric based on distributional measures, following MixSim~\cite{suo2023mixsim} and other prior work~\cite{montali2024waymo, zhong2023language}. 
In particular, we utilize the Wasserstein distance (WD) over adversarial ``profiles''---normalized histograms constructed from the adversary's yaw rates, acceleration values, and out-of-road rates. All WD values are computed via comparison to profiles derived from the original $X_{\text{adv}}$ in $\mathbf{S}$, which we average to compute an overall \texttt{Realism} meta-metric.\textcolor{red}{\textbf{[Reviewer 4, Item 5]:}} \textcolor{purple}{We also report relative collision velocities along the contact normal, as in \cite{rempe2022generating}, along with head-on collision rates and severe head-on rates (where severity is defined as collision velocity exceeding 5 m/s, thereby filtering out low-speed, glancing incidents).}

\begin{figure*}[ht!]
    \centering
    \includegraphics[width=0.92\linewidth]{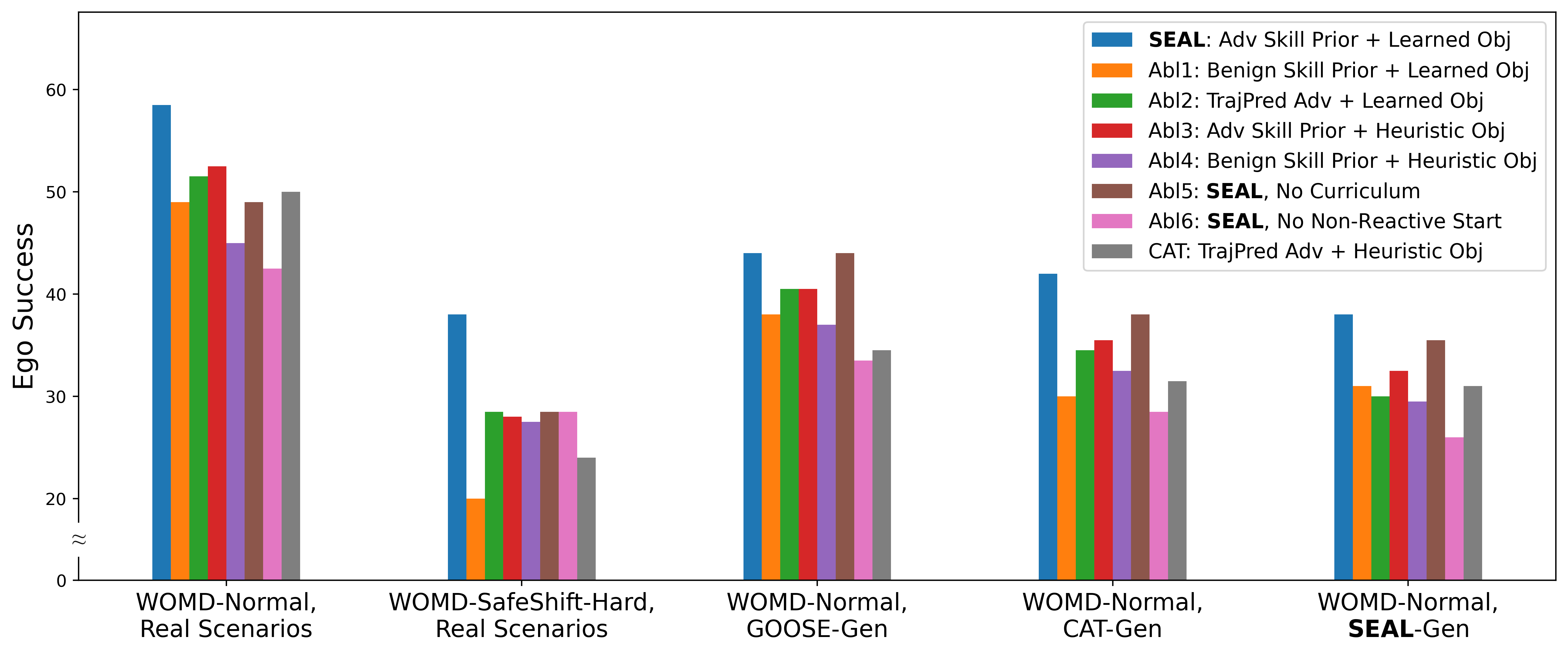}
    \vspace{-0.2cm}
    \caption{Ablation study on SEAL scenario generation training pipelines. Our full approach with learned objectives (\Cref{ssec:learned_obj}) and adversarial skill policies (\Cref{ssec:adv_skill}) produces the strongest downstream agents, across all five evaluation settings.}
    \label{fig:ablation}
    \vspace{-0.5cm}
\end{figure*}

\begin{figure}[t]
    \vspace{0.2cm}
    \centering
    \begin{subfigure}{0.24\textwidth}
        \includegraphics[width=\textwidth, page=1]{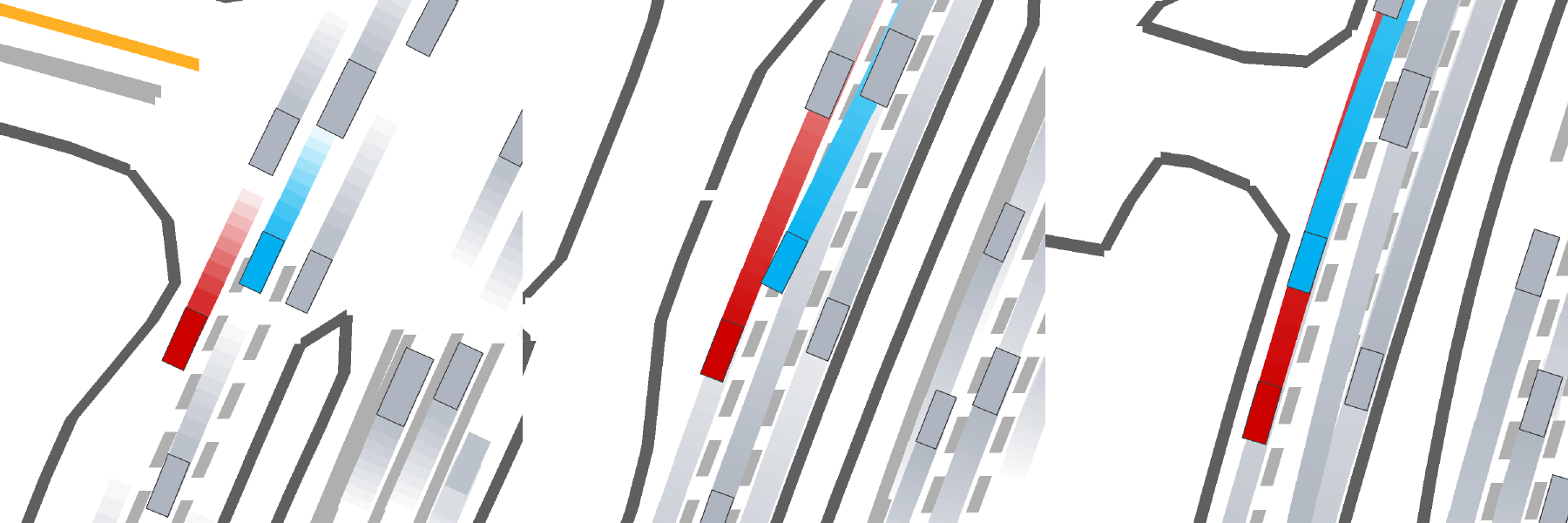}
        \caption{\textcolor{purple}{Ego Replay}}
    \end{subfigure}
    \hspace*{\fill}
    \begin{subfigure}{0.23\textwidth}
        \includegraphics[width=\textwidth, page=2]{figures/SEAL_qual_figures_ego.pdf}
        \caption{\textcolor{purple}{GOOSE~\cite{ransiek2024goose}-Trained}}
    \end{subfigure}
    \begin{subfigure}{0.23\textwidth}
        \includegraphics[width=\textwidth, page=3]{figures/SEAL_qual_figures_ego.pdf}
        \caption{\textcolor{purple}{CAT~\cite{zhang2023cat}-Trained}}
    \end{subfigure}
    \hspace*{\fill}
    \begin{subfigure}{0.23\textwidth}
        \includegraphics[width=\textwidth, page=4]{figures/SEAL_qual_figures_ego.pdf}
        \caption{\textcolor{purple}{\textbf{SEAL}-Trained}}
    \end{subfigure}
\caption{\textcolor{red}{\textbf{[Reviewer 2, Item 3]: }}\textcolor{purple}{Qualitative examples of \textbf{driving policies}; the \textcolor{egoblue}{\textbf{blue}} ego is a learned agent while the \textcolor{advred}{\textbf{red}} adversary is fixed. (a) shows the original human trajectory from \texttt{WOMD-SafeShift-Hard}, while (b), (c), and (d) show ego behaviors learned in different pipelines.}}
\label{fig:ego_qualitative}
\vspace{-0.3cm}
\end{figure}

\section{Results} \label{sec:results}
\textcolor{red}{\textbf{[Reviewer 1, Item 5]: }}\textcolor{purple}{We report the median and interquartile range (IQR) over four seeds, for greater statistical robustness.} These statistical summaries are computed independently over each metric, so \texttt{Success}, \texttt{Crash}, and \texttt{Out of Road} may not sum to $100\%$.  We also evaluate a non-reactive ego replay policy (\texttt{Replay}), which rolls out the original $X_{\text{ego}}$ trajectory, as well as a ReSkill \cite{rana2023residual} agent trained without any adversarial scenario generation (\texttt{No Adv}). Note that due to re-simulation limitations, \texttt{Replay} in \texttt{WOMD-Normal} and \texttt{WOMD-SafeShift-Hard} may have a nonzero failure rate. 

\noindent\textbf{Downstream Performance.} Our closed-loop training results are summarized in \Cref{tab:results_main}. SEAL-trained policies average a $\mathbf{21.5\%}$ \textbf{increase} in \texttt{Success} rate relative to the top baseline in each setting, achieving a strong balance between \texttt{Crash} and \texttt{Out of Road} rates. While a baseline-trained policy may have slightly better performance on one failure type, it is achieved by sacrificing performance against the other. 
\textcolor{red}{\textbf{[Reviewer 2, Item 2]: }}\textcolor{purple}{Compared to GOOSE and CAT, SEAL training yields more realistic yaw and road compliance but less realistic acceleration, indicating stronger braking and more disciplined in-lane maneuvering to manage criticality. Despite high kinematic realism, \texttt{No Adv} egos crash frequently due to lack of experience in challenging scenarios and resulting poor reactivity.}

\textcolor{red}{\textbf{[Reviewer 2, Item 3]: }}\textcolor{purple}{We highlight qualitative examples of ego behavior in \Cref{fig:ego_qualitative}, showcasing how different training regimes influence the execution of the same \textit{benign} skills, in a scenario drawn from the \texttt{WOMD-SafeShift-Hard} set of real, safety-relevant scenarios. While all ego policies operate within the same offline-learned skill space, their online adaptation differs across training methods. The \texttt{Replay} ego depicts the ground-truth human trajectory, which merges safely into the right lane. The GOOSE-trained ego initiates the merge too early and fails to recover in time, resulting in a collision. The CAT-trained ego begins to merge later but hesitates under pressure, slows down, and is rear-ended. In contrast, the SEAL-trained ego merges with a sufficient gap while accommodating a close-following tail vehicle, resulting in a smooth and safe maneuver.} \textcolor{blue}{These differences highlight how SEAL's more realistic and nuanced adversarial training scenarios better prepare ego policies to navigate challenging interactions effectively.}

\begin{figure}[t]
    \vspace{0.2cm}
    \centering
    \begin{subfigure}{0.24\textwidth}
        \includegraphics[width=\textwidth, page=1]{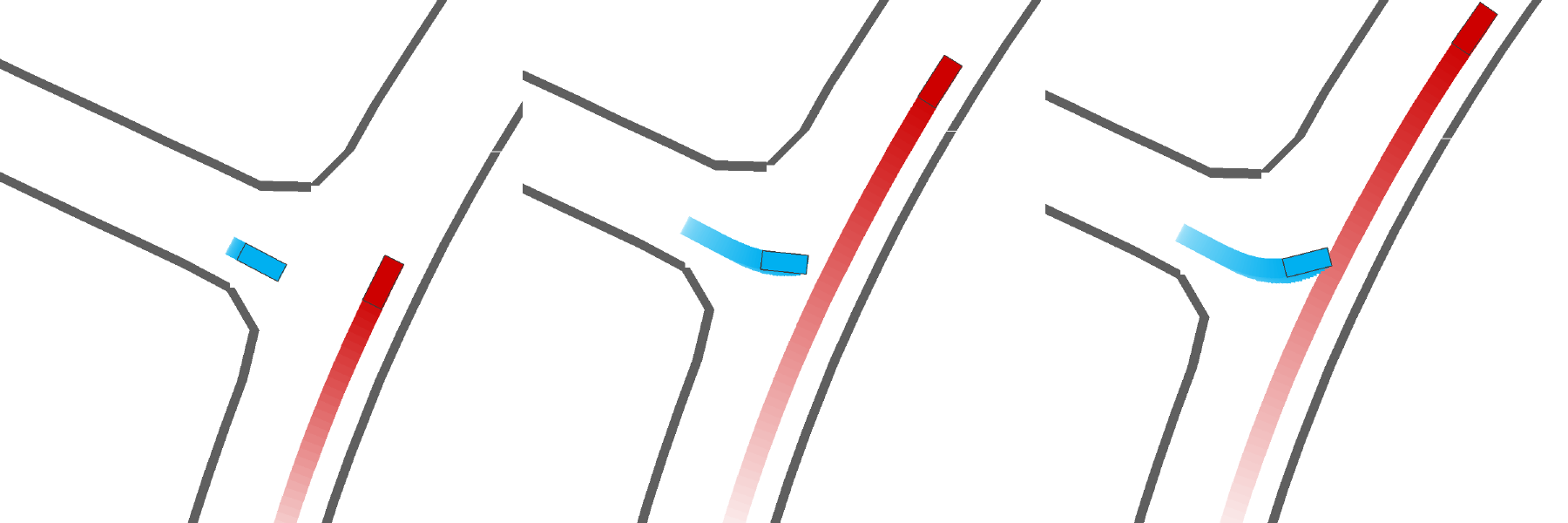}
        \caption{\textcolor{blue}{No Perturbation}}
    \end{subfigure}
    \hspace*{\fill}
    \begin{subfigure}{0.23\textwidth}
        \includegraphics[width=\textwidth, page=2]{figures/SEAL_qual_figures.pdf}
        \caption{GOOSE~\cite{ransiek2024goose}-Gen}
    \end{subfigure}
    \begin{subfigure}{0.23\textwidth}
        \includegraphics[width=\textwidth, page=3]{figures/SEAL_qual_figures.pdf}
        \caption{CAT~\cite{zhang2023cat}-Gen}
    \end{subfigure}
    \hspace*{\fill}
    \begin{subfigure}{0.23\textwidth}
        \includegraphics[width=\textwidth, page=4]{figures/SEAL_qual_figures.pdf}
        \caption{\textbf{SEAL}-Gen}
    \end{subfigure}
\caption{\textcolor{blue}{Qualitative examples of \textbf{scenario perturbation}; the \textcolor{egoblue}{\textbf{blue}} ego follows a fixed replay policy while the \textcolor{advred}{\textbf{red}} adversary is modified. (a) shows the original \texttt{WOMD-Normal} scenario, while (b), (c), and (d) show perturbations generated by GOOSE, CAT, and SEAL, respectively.}}
\label{fig:qualitative}
\vspace{-0.5cm}
\end{figure}

\noindent\textbf{Scenario Generation Quality.} %
\label{ssec:scene_gen_results} %
To directly assess scenario generation quality, we aggregate metrics in \Cref{tab:gen_quality}, averaged over all ego methods.
\textcolor{red}{\textbf{[Reviewer 4, Item 5]: }}\textcolor{purple}{Although CAT scenarios induce a lower ego \texttt{Success} rate and raw head-on rate than SEAL scenarios}, SEAL scenarios exhibit the highest \texttt{Realism} among scenario generation approaches, a $\mathbf{35.3\%}$ \textbf{improvement}, contributing to SEAL-trained policies' superior downstream performance.\textcolor{red}{\textbf{[Reviewer 4, Item 5]: }}\textcolor{purple}{Furthermore, SEAL's collision velocities and severe head-on rates are far lower than baseline approaches.}

\textcolor{red}{\textbf{[Reviewer 2, Item 3]: }}\textcolor{purple}{We also showcase qualitative examples of the tested scenario generation approaches in \Cref{fig:qualitative}, using a fixed ego \texttt{Replay} policy to isolate differences in adversary behavior. CAT and GOOSE both produce aggressive trajectories that lead to collisions: CAT stops and turns directly into the ego, while GOOSE swerves across the lane and slows in the ego's path to force a t-bone. In contrast, the SEAL adversary exhibits more nuanced behavior, slowing down to let the ego catch up, moving away at the last moment to induce a near-miss, and thus demonstrating interesting \textit{adversarial} skill behavior.}

\noindent\textbf{Ablation Studies.} %
\label{ssec:ablation} %
To further investigate how different components of SEAL affect downstream training, we perform extensive ablation studies shown in \Cref{fig:ablation}, as well as comparing against CAT as it is a slightly stronger baseline than GOOSE. We study the effect of our learned objective function by comparing it to the heuristic, bounding box overlap approach used by CAT (\texttt{Learned Obj} and \texttt{Heuristic Obj}, respectively). Similarly, we compare our adversarial skill policy (\texttt{Adv Skill Prior}) with a benign prior variant (\texttt{Benign Skill Prior}) and a predefined trajectory following policy (\texttt{TrajPred Adv}).
\textcolor{red}{\textbf{[Reviewer 4, Item 3]: }}\textcolor{purple}{We also compare SEAL against two additional ablations: one trained \textit{without} curriculum, and another \textit{without} the initial non-reactive start.
\textcolor{black}{Our full SEAL approach performs best across all settings; both the learned objective function and adversarial skill policy are essential}, while the curriculum and non-reactive start further improve performance.}

\section{Conclusion} \label{sec:conclusion}

As autonomous driving (AD) systems advance, ensuring safety remains essential. While recent safety-critical scenario generation techniques show promise, they often lack the realism, reactivity, and nuance needed to provide strong training signals for closed-loop agents. We thus introduced Skill-Enabled Adversary Learning (SEAL) as a perturbation-based safety-critical scenario generation approach, combining a learned objective function and an adversarial skill policy. In all test settings---across both real-world challenging scenarios and generated scenarios by SEAL and other SOTA methods---SEAL-trained policies achieved significantly higher success rates, with a more than 20\% relative increase. Upon deeper analysis, SEAL-generated \textcolor{purple}{scenarios} contain less aggressive but more realistic adversaries, helping to explain the observed ego agent improvements. We argue that realism metrics, downstream task utility, and out-of-distribution evaluation settings are vital in assessing adversarially-perturbed scenarios.

While SEAL is quite effective, further improvements are still possible. Incorporating finer-grained metrics into the objective function could enable more adaptive and controllable generation beyond safety criticality alone. Additionally, enhancing realism metrics to reflect human decision-making at the skill-level could provide deeper insights into scenario quality. We encourage future work to explore these topics.


\addtolength{\textheight}{0cm}   



\section*{ACKNOWLEDGMENT}
This work was partially performed during Benjamin Stoler's internship at Stack AV; the authors thank Stack for their mentorship. The authors additionally thank Evan Lohn for many valuable discussions throughout the design process. 





\bibliographystyle{IEEEtran}
\bstctlcite{IEEEexample:BSTcontrol}
\bibliography{ref}

\end{document}